\RequirePackage{fix-cm}
\documentclass[smallextended]{svjour3}       
\smartqed  
\usepackage{graphicx}
\usepackage{comment}
\usepackage{textcomp}
\usepackage{subcaption}
\usepackage{amsmath}
\usepackage{amsbsy}  
\usepackage{tabularx} 
\usepackage{adjustbox}
\usepackage{tikz}
\usetikzlibrary{shapes, arrows, positioning, calc}

\graphicspath{ {./images/} }
\begin{document}

\title{Improved colonoscopy polyp detection with massive endoscopic images
}


\author{Jialin Yu         \and
        Huogen Wang \and
        Ming Chen    
}


\institute{Jialin. Yu \at
              HiThink Royalflush \\
              \email{ucabjj2@ucl.ac.uk}           
}
 
\date{Received: date / Accepted: date}

\maketitle 

\begin{abstract}

We improved an existing end-to-end polyp detection model with better average precision validated by different dataset with trivial cost on detection speed.
Previous work on detecting polyps within colonoscopy \cite{Chen2018} provided an efficient end-to-end solution to alleviate 
doctor's examination overhead. However, our later experiments found this framework is not as robust as before as the condition of polyp capturing varies. 
In this work, we investigated dataset, identifying main issues that causes low precision rate in the task of polyp detection. We used an optimized anchor generation methods
to get better anchor box shape and more boxes are used for detection as we believe this is necessary for small object detection. An alternative backbone is used to compensate the heavy 
time cost introduced by dense anchor box regression. With use of the attention gate module, our model can achieve state-of-the-art polyp detection performance while still maintain real-time detection speed.

\end{abstract}

\section{Introduction}
\label{intro}
A 2020 study \cite{Siegel2020} reports that mortality of colorectal cancer(CRC) has become the second most in US. And the global increasing trends of 
CRC might grow continuously over next decade and it is predicted to introduce 2.2 million new incidences by 2030 \cite{Arnold2016}. 
Nevertheless, figures in highly indexed HDI countries is relatively lower than those with medium HDI, partially due to the application of 
polypectomy \cite{Arnold2016,Center2009}. 

Polys are some abnormal cells resides the intestine from colon surface, and are likely to develop into cancer tumors through time without
any treatment \cite{Sundaram2019}. Polypectomy is a surgical process of excision precancerous polys after their presence is detected. Therefore, 
as many researches \cite{Fauzi2019,Sundaram2019} suggested, early detection of polyps is important to prevent colon cancer. To date, colonoscopy or endoscopy
are used to examine polyps condition from a patient's colon, consequently, medical imagery are growing rapidly with the increasing popularity of 
these screening methods. However, doctors can not directly utilizing the abundance in existing visual dataset in increasing their diagnose ability, whereas 
more manual examination overheads are introduced. To this end, a fully automated detection system can be suggested to highlight potential region of interest.  For this reason, an 
end-to-end real-time polyp detection system \cite{Chen2018}  which can be used to alleviate
doctors' cost while maintaining a stable detection performance. This proposed system features a end-to-end training pipeline with good performance on minor polyp object detection thanks to the adaption of 
RetinaNet \cite{Lin2017} as a detector. However, RetinaNet is designed to be concise in order to highlights the performance gain comes solely from authors
proposed focal loss and their underlying concepts. 

\begin{figure}
    \centering
    
    \begin{subfigure}[t]{.4\textwidth}
    \centering
    \includegraphics[width=\linewidth,height = 4.5cm]{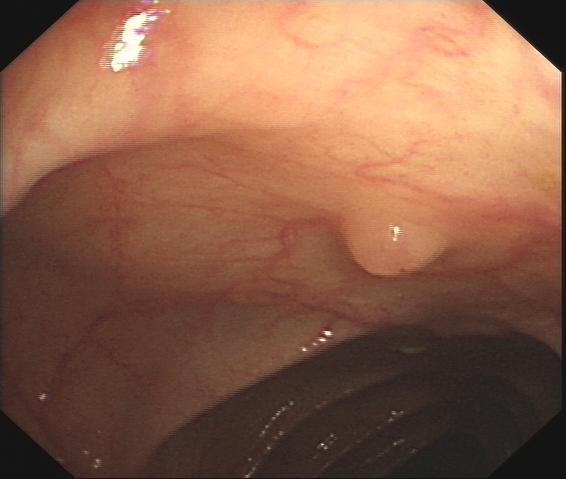}
            \caption{sessile}\label{fig:fig_a}
    \end{subfigure}
    \begin{subfigure}[t]{.4\textwidth}
    \centering
    \includegraphics[width=\linewidth]{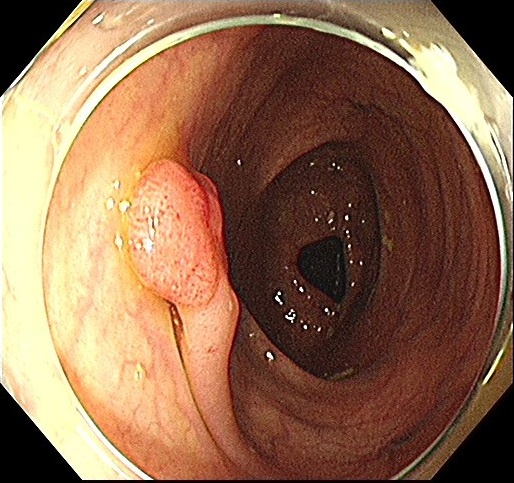}
    \caption{pedunculated}\label{fig:fig_b}
    \end{subfigure}
    
    \medskip
    
    \begin{subfigure}[t]{.4\textwidth}
    \centering
    \vspace{0pt}
    \includegraphics[width=\linewidth]{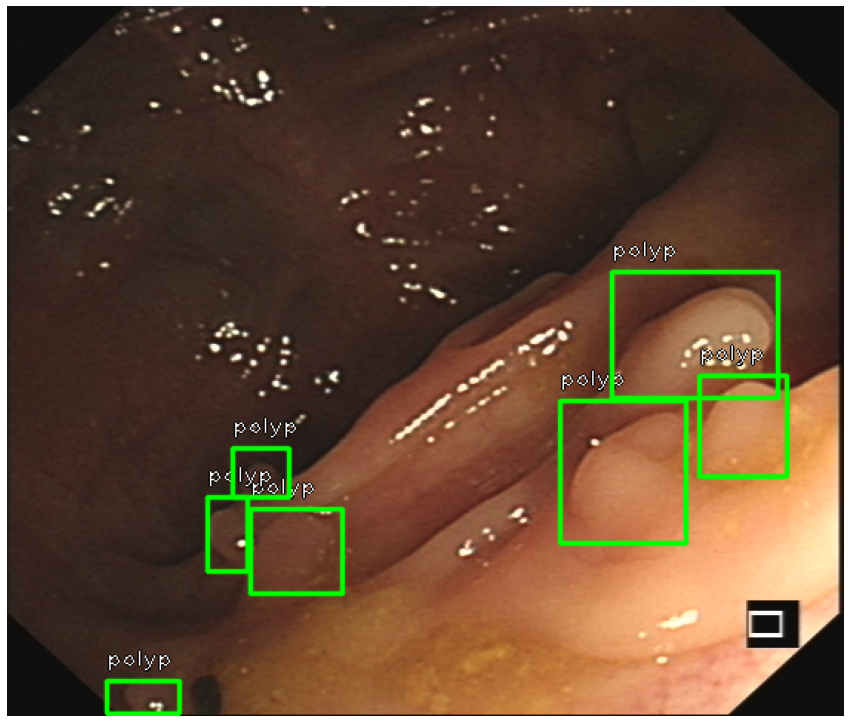}
    \caption{multiple objects}\label{fig:fig_c}
    \end{subfigure}
    \begin{minipage}[t]{.4\textwidth}
    \caption{Examples of possible polyp category within a frame: (a) sessile, (b) pedunculated or (c) multiples.}\label{polype}
    \end{minipage}

    \end{figure}

General objects for detection task such as car, pedestrian or table which have defined outline or appearance which offers strong cues for neural network to learn. Polyps, on the other hand, 
have various sizes which scales from diminutive($<5mm$) to giant($>30mm$) \cite{Shussman2014} and present different appearances such as sessile or pedunculated \figurename{\ref{polype}}. This makes detecting polyps using a general-purpose oriented 
network more difficult and urges us to design a object-specific detector.

For this work, we collected two more datasets for testing with one of them being significantly more challengeable.  
According to our empirical test of RetinaNet with our new test sets, we find the conciseness of such network design could result in a unreliable detection rate. We revisit the potential causes for 
performance degradation and some improvements has been proposed to maintain a comparable result when difficult sample is given. 
We propose new backbone and subnet architectural designs by incorporating dilated convolution, this allows network having larger receptive field with better multi-scale performance
 without increasing depth. To enhance the network in detecting various scale object, we also optimize the anchor configuration based on heuristic search and employ a attention gated function during feature fusion stage. 
Additionally, we further improved our detector by using different post-processing mechanism the Soft-NMS \cite{Bodla2017}. 

\section{Object Detection}
\subsection{Multi-stage object detection} 
An object detection task aims at locating where an instance of object may exists. Detection algorithm predates the heavily use of deep learning era usually be referred as traditional detector. Many hand-crafted
descriptors such as HOG,SIFT are proposed to extract features over a sliding-window mechanism. The introduction of DPMs and its variants push the sliding-window
 to its peak by dominating PASCAL competition for many years. The throne of sliding-window based detectors are transferred after the resurgence of Deep Learning, where CNN based feature extraction shows prominent 
 results. CNN based object detectors consists of two parts, feature extractor and regressor. Depending on how potential bounding box are generated, detectors will also be classified as one-stage or two-stage.

The two-stage paradigm can be simply summarized as set of candidate proposals are generated during the first stage, these proposals are regions that network believes to have object of interest. The second stage 
will refine these proposals to classify them into background and foreground.  Selective Search network \cite{Uijlings2013} first proposed this idea and it is surpassed by R-CNN\cite{Girshick2014} with the change of 
having CNN as the proposal classifier. R-CNN went through several upgrades in terms of both speed and accuracy. Fast R-CNN\cite{Girshick2015} uses feature map to generate region of proposals and  
RoI pooling to fix the input size, thus less convolution operations are required per image. Faster R-CNN\cite{Ren2017} makes processing image faster by replacing the time-consuming selective search with 
Region Proposal Networks. Extensions in this paradigm based on faster R-CNN framework can be found in \cite{Zhong2020}\cite{Shrivastava2016}\cite{Shrivastava2016a} with focus on achieving higher accuracy. However, 
all these two-stage detectors are hardly meet the standard of real-time for video sequence. 

Compared to proposal-based object detection network, single stage detector achieves higher frames per second(fps) with cost of accuracy. One stage detector uses anchor to replace the use of proposal.
Anchor boxes differ proposals in a way of how they are generated, each box have predefined ratio and scale. Thus, anchor design introduce author's prior toward the task they are addressing. 
Performance of early design of single-stage detector differs much to two-stage with focus on achieving higher speed with lost of accuracy. SSD \cite{Liu2016ssd} and YOLO \cite{Redmon2016} are typical instance
 in this era. One known issue barricading these networks achieving higher accuracy is the so-called class-imbalance problem where the classifier requires to process numerous background images but limited foreground images. This problem is studied in \cite{Lin2017} with a feasible solution of using focal loss, it uses specific coefficients 
 to prevent backgrounds overwhelms the classifier.
Successors of single shot networks, such as EfficientDet and YOLOv4, both demonstrates comparable performance to those two-stage network while maintaining a optimal speed.These state-of-the-art designs either employ
focal loss directly or creating dedicated structure to handle the class-imbalance problem. 

\subsection{Existing Polyp Detection Framework}

Conventional polyp detection works uses engineered features such as color wavelet \cite{Karkanis2003}, elliptical shape\cite{Hwang2007}, edge \cite{Wang2014}, valley and contour \cite{Bernal2012} to classify the 
presence of object within a given region. However, many hand-crafted descriptors are proposed to address specific detection issus thus, failed to be robust to variations of object in shapes, size, illumination etc.

In contrast to conventional approach, convolutional neural network gives unparalleled feature generalization ability with sufficient data availability makes CNN an ideal feature extractor for
polyp detection task. Initial attempts of incorporating CNN into polyp detection task \cite{Tajbakhsh2015} uses engineering feature as candidate region, ensembled CNNs are used to extract robust feature out of 
polyps within these manual regions. In medical video domain, a 3D-CNN is favored by some researchers as it learns temporal information whereas 2d convolution only perceives spatial data. \cite{Yu2017} proposed a
3D-FCN polyp detection framework, the learned spatio-temporal features from 3D convolution successfully reduced the variations of intra-class and intro-class within polyps. Convolution in 3D is 
essentially costly operation, though authors in \cite{Yu2017} demonstrates their efforts in acceleration, structures like this are hardly a plausible choice for a real-time detection system. All these discussed 
polyp detectors above are applying non-linear function to feature map to generate a probability map which is thresholded to get precise location of object. In this regard, \cite{Zhang2017} propose to use SVM as 
a classifier on to a transfer-learned feature map to predict the presence of a polyp. While those anchor-based or proposals-based framework have achieved noticeable gains in object detection, most works aim at achieving
higher results in finding general objects. Recently, researchers begin to adopt these popular framework into polyp analysis. \cite{BlanesVidal2019} uses jointly AlexNet and R-CNN for their autonomous polyp detection
algorithm where they add additional pooling layer and use different optimization function to avoid overfitting issue. Three types of  pooling layers --namely second-max pooling, second-min pooling
and min-pooling -- are added to SSD to improve detection result for small polyp \cite{Zhang2019}.  Authors from a 2020 study \cite{Jia2020} propose a two-stage polyp localization framework where the faster R-CNN is used to generate polyp proposals followed by a polyp segmentation
stage. Semantics in both stages are shared thus assure the model's performance.

Previous work \cite{Chen2018} provides a end-to-end method, as given in \figurename{\ref{fig:pipeline}, for detecting polyps where the input of colonoscopy content are feed directly 
into the system and precise location of polyps will then be presented as results.The overall pipeline of such system are organized as follows: an image data acquisition card is used to capture endoscope data and all data
will be decoded into a RGB video sequence. The original input sequence consists of some undesired content: polyp-irrelevant and non-colorectal. Two trained  pre-process network \cite{Chen2018} is used to remove these
elements.  

\begin{figure} 
    \includegraphics[width=0.9\linewidth]{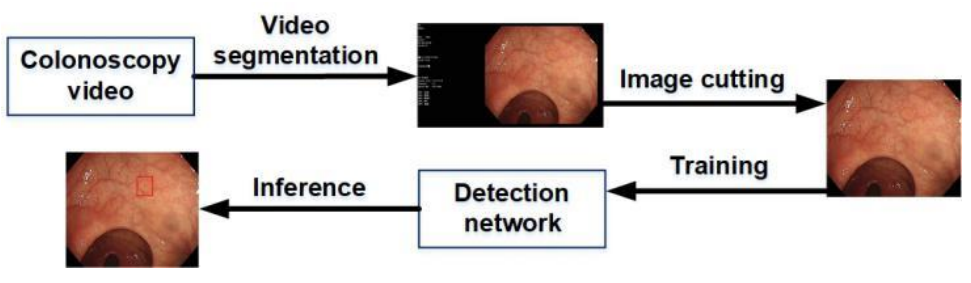}
  \caption{Overall pipeline demonstration for polyp detection}
  \label{fig:pipeline}       
  \end{figure}
Non-colorectal data is generated by improper use of endoscope where an 
operator does not start or end the acquisition at the appropriate time, therefore, results in unnecessary non-colorectal content being recorded. Author trained a classification network to differentiate the presence condition
(internal or external) for each frame then follows a thresholding mechanism to smooth the inconsistency within consecutive frames. The output of threshold consist inside-body clip only and will be further pre-processed namely
cutting, to remove the text box and other non-colorectal. The cutting process employs a manually-tuned binary mask representing the largest connected-component to preserve colorectal-only images. This will form the 
input data stream to our detection network. 

The original method incorporate single stage RetinaNet\cite{Lin2017} as a detector. Although detecting polyp is a single object detection task, authors trained network with multiple classes where each category comes from 
different false positive instance. This indeed prone to work better but only in pre-defined situations since network trained in this way are based on assumption that these defined classes are diverse enough to 
cover any class of object within colon.

\section{Methods}
An overview of our proposed polyp detector is given in \figurename{\ref{fig:architecture}}. We first show how our network architecture is designed including FPN and anchor configuration. We then describe other components which benefits in achieving higher
detection accuracy.
\subsection{Network design}
RetinaNet is a famous one-stage object detector, it uses focal loss to address the common detection problem called class imbalance. Additionally, RetinaNet adopts a top-down Feature Pyramid Network 
(FPN) from \cite{Lin2016}
which extended a CNN with lateral connections to construct a pyramidal multi-scale feature architecture. Our baseline will follow the structure of RetinaNet, study of choosing optimal backbones ResNet-50 over VGG are 
discussed in \cite{Chen2018}.  
The original structure of FPN used by RetinaNet consists of 5 stages from $P_3$ to $P_7$, each level can be used to detect object in different scales. Concretely, level $P_3$ to $P_5$ are obtained from the shortcut
output of applying top-down and lateral connection to ResNet blocks $C_3$ to $C_5$. $P_6$ is computed by 2-strided convolution on $P_5$ and $P_7$ is a down-sampled $P_6$ activated by ReLu. Coarser level $P_6$ and $P_7$
which use for large object detection is computed based on $P_5$ instead of feature learned by ResNet. Such design speeds up computation with trivial lost on detecting larger object. Our design, on the other hand, 
uses different method which will be discussed later to improve speed, therefore, all 5 levels will be computed in the same way as in \cite{Lin2016}.  
\begin{figure}[h] 
    \includegraphics[width=\linewidth]{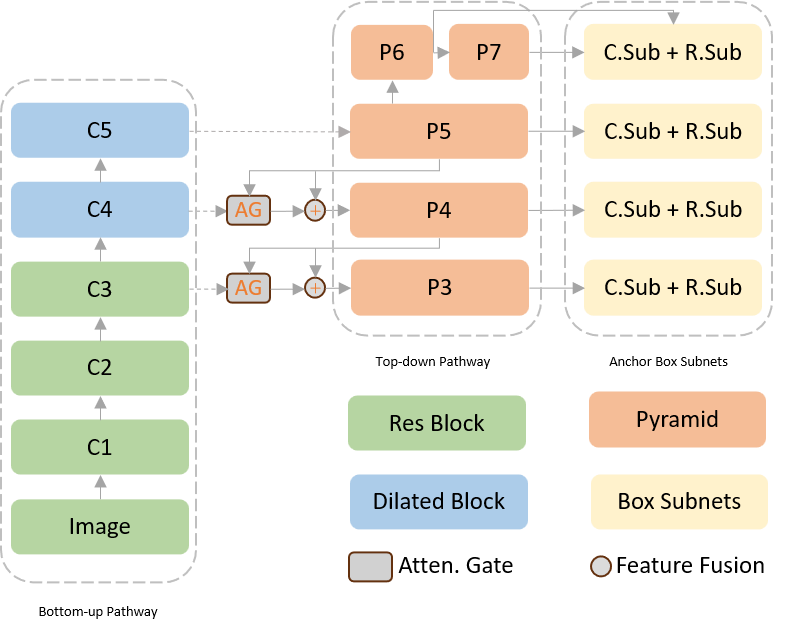}
  \caption{An illustration of our architecture design: An image will first go to Bottom-up Pathway(Backbone) for feature map accumulation, a Feature Pyramid Network is build upon on the output of Backbone as well as the residual blocks.
  All stages(P3 to P7) from feature pyramid will be futher shared by two subnets.}
  \label{fig:architecture}       
  \end{figure} 
 \subsection{Anchor optimization}
Anchors with different sizes and rations are generated on top of each pyramid level and feed into two subnets which are used to classify label and regress bounding box. Original anchor configurations uses sizes (32, 
64, 128, 256, and 512) each size has three scales $(2^{0}, 2^{\frac{1}{3}}, 2^{\frac{2}{3}})$ and three aspect ratios $(1:2, 1:1, 2:1)$. As the quality of anchors will have direct impact on detection,
keeping default anchor settings might not possible to detect multiple polyps smaller than 32 within a frame. In this regard, we use differential evolution algorithm \cite{Storn1997} to approximate optimal anchor
configurations iteratively. Process of optimizing anchor configurations are detailed in \cite{Zlocha2019a}, an object function which maximizes overlap area between anchor and object bounding box are optimized by
 a group of iteratively improved candidate solutions. We increase the number of anchors per level from 9 to 15 by having 5 aspect ratios so that each pyramid scale has denser anchor coverage. 
 Each scale in our method is optimized using its corresponding stride value. We use validation set to find optimal configurations, we eventually find the following anchor parameter combinations with scales$(16\: 32\: 64\: 64\: 64)$ and ratios $(0.481\: 0.741\: 1.0 \:1.349\: 2.078)$ 
fit our task best.

The \figurename{\ref{fig:anchoroptimize}} depicts the effect of anchor optimization. All boxes are centering at the object, after applying the optimized anchor parameters, each objects are associated with more boxes which also 
cover more area of polyps. 
  \begin{figure}[h]
    
    \begin{subfigure}{0.5\textwidth}
    \includegraphics[width=0.9\linewidth, height=5cm]{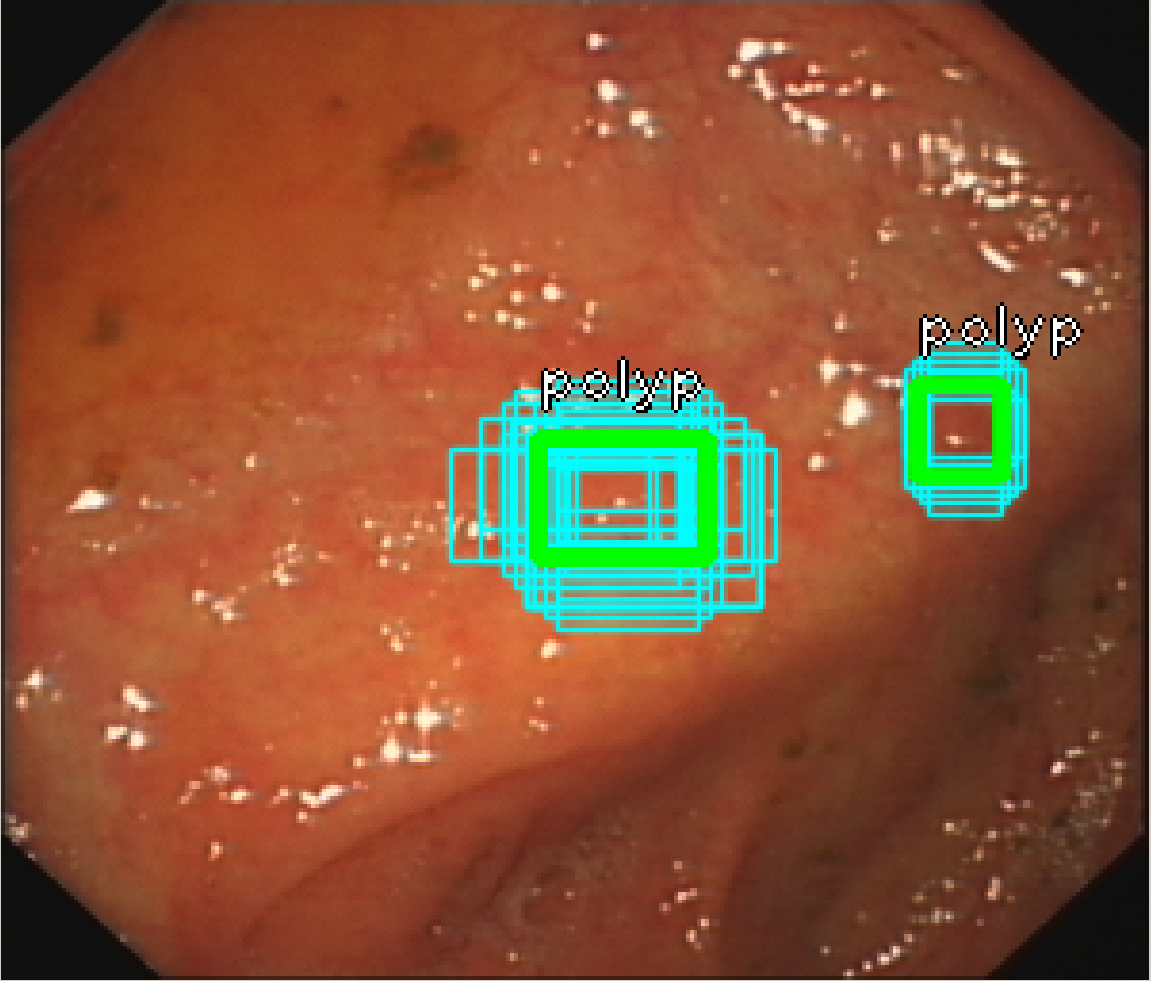} 
    \caption{sparse anchor coverage}
    \label{fig:sparse}
    \end{subfigure}
    \begin{subfigure}{0.5\textwidth}
    \includegraphics[width=0.9\linewidth, height=5cm]{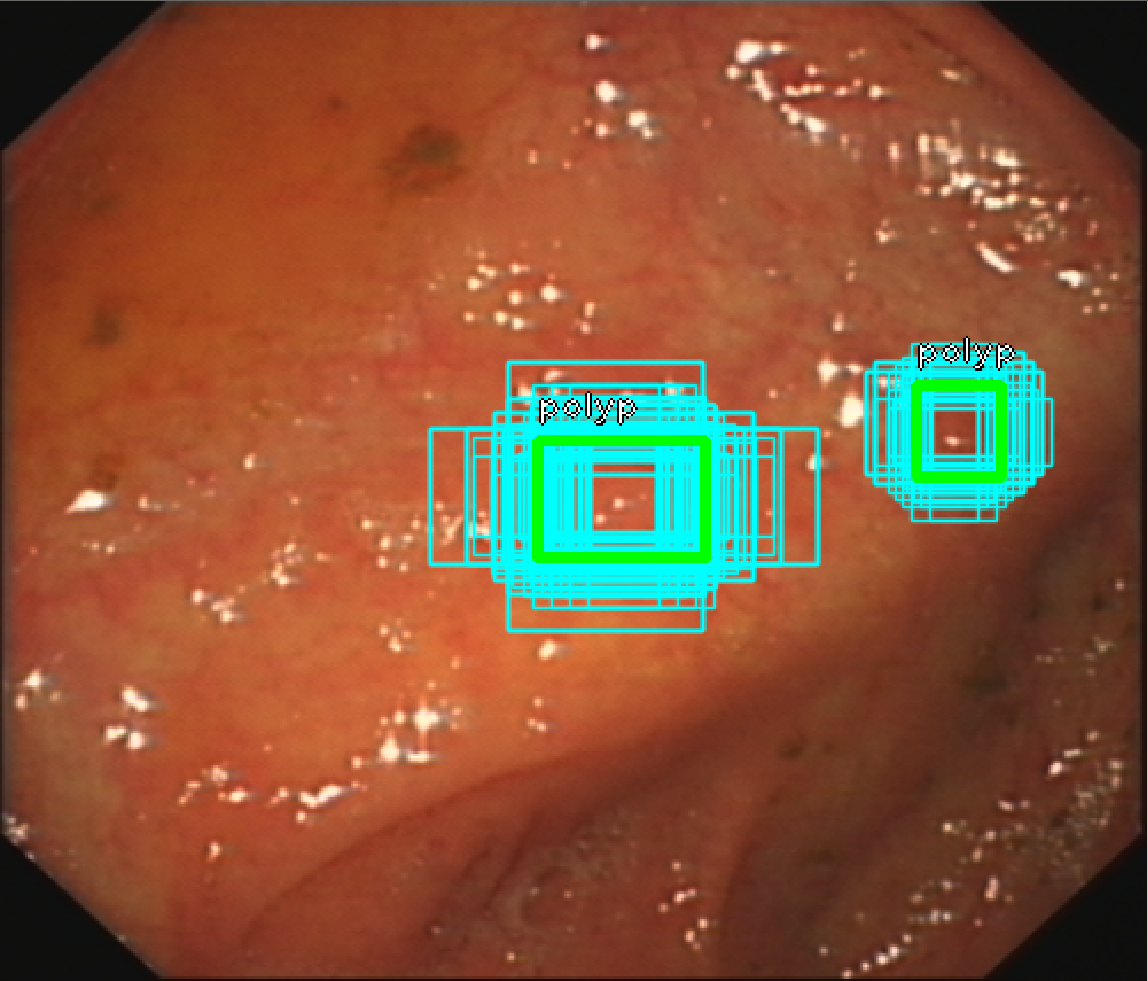}
    \caption{dense anchor coverage}
    \label{fig:dense}
    \end{subfigure}
    \caption{example of anchor optimization. Anchor boxes on the left are the result of optimization, the ratio and sizes offer better coverage 
    with GT label and more boxes are used.}
    \label{fig:anchoroptimize}
    \end{figure}

\subsection{Dilated Convolution}
Previous sections briefly discussed how we get optimal anchor configurations, we increase the number of anchors for each pyramid scale to increase multi-object and finer object detection performance(boundary anchor).
More anchor introduce heavier computation complexity which necessitate a efficient backbone design to even out the load from anchors. A traditional convolutional neural network reduces the resolution but increase the 
receptive field progressively. However, the very last feature map from CNN might be too coarse to be discernible. This issue is addressed by Dilated Residual Network (DRN)\cite{Yu2017a}. A dilated convolution, also known
as atrous convolution, is defined as \cite{Yu2015}:
\begin{equation}
\left(F *_{l} k\right)(\mathbf{p})=\sum_{\mathbf{s}+l \mathbf{t}=\mathbf{p}} F(\mathbf{s}) k(\mathbf{t})
\end{equation}
where $l$ is the dilation rate. This operation introduces a space between values within the kernel (i.e., if we dilated a 3 by 3 convolution with rate of 2, then the resulting receptive filed size is equivalent to the ones 
of 5 by 5 convolution). One advantage of using dilation is that we can achieve same receptive field size but higher resolution with less parameters compare to its non-dilated counterparts. Thus, as reported in \cite{Yu2017a}
,dilating the convolution layer can, to some extent, increase the performance without the need of increasing depth or complexity.  
\begin{figure}[h] 
    \includegraphics[width=0.8\linewidth]{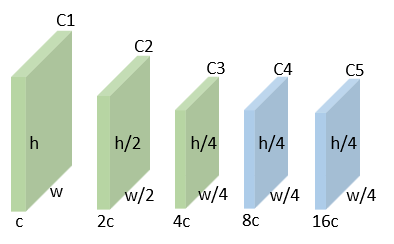}
  \caption{Layer resolution with dilated convolution.}
  \label{fig:dilate size}       
  \end{figure} 

Nevertheless, dilation is known to have gridding effect, an artifacts which occurs when sampling rate of dilation is lower than the frequency in feature map. As shown in \figurename{\ref{fig:gridded}}, a dilation with rate
2 could make output with a grided pattern. We adopt some preliminary degridding procedures as studied in\cite{Yu2017a} \cite{Triggsa} by replacing the first global pooling layer with ResNet block and adding extra residual blocks with gradually decreasing
dilation rate convolution. The overall structural design of our backbone is inspired by DetNet \cite{Li2018}, we replace the original bottleneck block $C_4$ and $C_5$ from ResNet-50 with the dilated bottleneck and 1x1 conv projection. 
The overall refined backbone \figurename{\ref{fig:gridded}} shows a clear transition in layer resolutions, where the integration of dilated convolution progressively restores finer spatial details across multiple scales.
On top of that, we applied the degridding mechanism to above structure and $P_6$ are now computed directly from backbone. \figurename{\ref{fig:gridding}} shows the de-gridded feature map.
  \begin{figure}[h]
    \begin{subfigure}{0.5\textwidth}
    \includegraphics[width=0.9\linewidth, height=5cm]{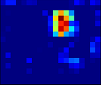} 
    \caption{gridded}
    \label{fig:gridded}
    \end{subfigure}
    \begin{subfigure}{0.5\textwidth}
    \includegraphics[width=0.9\linewidth, height=5cm]{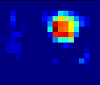}
    \caption{degridded}
    \label{fig:degridded}
    \end{subfigure}
    \caption{Example of (a)gridded and (b) degridded feature map. Heat value in degridded feature map gives smoother spreads compare to its gridded map.  }
    \label{fig:gridding}
  \end{figure}

\subsection{Attention gated block}
Applying attention gate to small and varying object has been proven to work according to \cite{Zlocha2019a}. It is a light-weight yet efficient accuracy-boosting mechanism, it only requires a 1x1 convolution 
to produce an attention map. This enables model finding feature saliency while suppressing irrelevant regions. Essentially, it is a learned coefficient $\alpha$ scaling the feature vector $x$, a global feature
vector $g$ from coarse spatial level is used to disambiguate non-task information for AG. The attention gate module are modelled by so-called additive attention as formulated by:

\begin{equation}
  \begin{array}{c}
    q_{a t t, i}^{l}=\boldsymbol{\psi}^{T}\left(\sigma_{1}\left(\boldsymbol{W}_{x}^{T} \boldsymbol{x}_{i}^{l}+\boldsymbol{W}_{g}^{T} \boldsymbol{g}+\boldsymbol{b}_{x g}\right)\right)+b_{\psi} \\
    \alpha^{l}=\sigma_{2}\left(q_{a t t}^{l}\left(\boldsymbol{x}^{l}, \boldsymbol{g} ; \boldsymbol{\Theta}_{a t t}\right)\right)
    \end{array}
\end{equation}

Practically, we use output from each pyramid level as the gate signal $g$ and output from skip connection as input feature vector $x$.

\section{Experiments and results}
For experiments, both training and testing are conducted on Nvidia GTX1080 Ti. Additionally, we use TensorFlow and Keras, the open-source deep learning frameworks to accompany our model's implementation.
\subsection{Datasets}
We annotated our colonoscopy dataset for training and evaluation. It consists of 80k images from colonoscopy capture card output and follows the pre-processing procedures, this 
results in different pixel resolution ranging from $431 \times 368$ to $751 \times 641$. All images which are verified by our 
experienced gastroenterologist either contains our polyp annotations or not. We design and conduct several experiments to evaluate detection performance and details of experiment and result will be given in the 
following sections.

\subsection{Training and inference}
We trained our network by following the strategies listed in \cite{Lin2017}. We use adam as optimizer with weight decay 0.0001 and momentum 0.9, a focal loss and $L_1$ loss are used for training and box regression.
Batch-size of 8 are used to for each epochs and group normalization with group size of 8 will be updated accordingly. All inputs are resized to 300x300 in accordance with size used in our previous work \cite{Chen2018}.
We also adopted flipping shearing for augmentation.

\subsection{Polyp Detection Results}
Our dataset is split into training, validation and testing with ratio of $(7:2:1)$. All networks are trained from scratch as there is no 
pre-trained weights for our proposed architecture. The performance of polyp detection is evaluated quantitatively by the following metrics as
shown in \ref{tab:metric}.
\begin{table}
  \centering
  \caption{Metrics for polyp detection}
\begin{tabular}{c}
  \hline  $\operatorname{Precision}=\frac{\mathrm{TP}}{\mathrm{TP}+\mathrm{FP}}$ \\[0.3cm]
  $\operatorname{Recall}=\frac{\mathrm{TP}}{\mathrm{TP} + \mathrm{FN}}$ \\[0.3cm]
  $\mathrm{F} 1=\frac{2 \times \operatorname{Prec} \times \mathrm{Rec}}{\mathrm{Prec}+\mathrm{Rec}}$ \\[0.3cm]
  $\mathrm{AP} =\int_{0}^{1} p(r) d r$ \\
  \hline
  \end{tabular}
  \label{tab:metric}
\end{table}
As a detection task, we use Intersection over union(IoU) to define a True Positive (TP) IoU of Ground truth(GT) and Prediction is greater than 0.5; otherwise, it is False Positive(FP).
A False Negative(FN) is defined as count of the absence of the detection of a frame. Precision (Prec) describes the proportion of positive return is correct, while recall(rec) describes 
the proportion of real positive output is predicted correctly. F1 score indicates the balance between prec and rec. The area under the precision-recall curve $p(r)$ is the Average Precision (AP).

\begin{table}
\begin{center}
  \caption{Polyp detection result.}
  \label{table:result}
\begin{tabular}{lccccc} 
  Method & \# of Pars & F1 & mAP & Recall &  Frame per Second \\
  \hline RetinaNet\cite{Chen2018} & $36.40 \mathrm{M}$ & $.821 $ & $.869 $ & $.814 $ &  $.0291 \mathrm{s}$ \\
  + AO  & $36.32 \mathrm{M}$ & $.825 $ & $.888 $ & $.833 $  & $.0422 \mathrm{s}$ \\
   + AO + AG & $36.39 \mathrm{M}$ & $.825 $ & $.891 $ & $.848 $  & $.0422 \mathrm{s}$ \\
   + AO + AG + DilatedConv & $30.77 \mathrm{M}$ & $.897 $ & $.866 $ & $.807 $  & $.0198 \mathrm{s}$ \\
   + AO + AG + DilatedConv + Degrid & $34.77 \mathrm{M}$ & $.901  $ & $.901 $ & $.807 $  & $.0222 \mathrm{s}$ \\
  \hline
  \end{tabular}
\end{center}

\end{table}
Table \ref{table:result} shows the polyp detection result on our test set with various component incorporated. For this result, the backbone of 
RetinaNet is ResNet-50. DilatedConv indicates layers from ResNet-50 are replaced by Dilated Conv. From this results, we can see the effect of adding each components. With the anchor optimization(AO) and Attention Gate(AG), 
the mAP increased by 2 percents but the processing speed drops significantly. We therefore adapted DilatedConv where less 
parameters are used without much performance drop. As discussed in the previous section, new backbone has a major issue also known as gridding effect. We observed that,
the fix of such effect with attention module gives significant improvement without noticeable penalties on detection speed.

\section{Conclusion and Future Work}
In this paper, we extends our previous studies on designing an End-to-End polyp detection model. We discussed the potential
reasons that cause our old model failed to give competitive results as before.
We also showed that the original RetinaNet with ResNet-50 backbone can be further improved using a   
task-specific anchor optimization strategy. We plant in a attention gate module and design a de-gridded backbone to 
guarantee a real-time detection speed. Impressive results have been given on polyp detection based on our private dataset which exhibit the efficiency and generality of our improved model.
Given the clinically available annotated dataset, the proposed method can be easily tuned and given fast inference, implying its applicability in many clinical practice.

\section*{Conflict of interest}
 The authors declare that they have no conflict of interest.


\bibliographystyle{spmpsci}      
\bibliography{main}   

\begin{thebibliography}{10}
\providecommand{\url}[1]{{#1}}
\providecommand{\urlprefix}{URL }
\expandafter\ifx\csname urlstyle\endcsname\relax
  \providecommand{\doi}[1]{DOI~\discretionary{}{}{}#1}\else
  \providecommand{\doi}{DOI~\discretionary{}{}{}\begingroup \urlstyle{rm}\Url}\fi

\bibitem{Arnold2016}
Arnold, M., Sierra, M.S., Laversanne, M., Soerjomataram, I., Jemal, A., Bray, F.: Global patterns and trends in colorectal cancer incidence and mortality.
\newblock Gut \textbf{66}(4), 683--691 (2016).
\newblock \doi{10.1136/gutjnl-2015-310912}

\bibitem{Bernal2012}
Bernal, J., S{\'{a}}nchez, J., Vilari{\~{n}}o, F.: Towards automatic polyp detection with a polyp appearance model.
\newblock Pattern Recognition \textbf{45}(9), 3166--3182 (2012).
\newblock \doi{10.1016/j.patcog.2012.03.002}

\bibitem{BlanesVidal2019}
Blanes-Vidal, V., Baatrup, G., Nadimi, E.S.: Addressing priority challenges in the detection and assessment of colorectal polyps from capsule endoscopy and colonoscopy in colorectal cancer screening using machine learning.
\newblock Acta Oncologica \textbf{58}(sup1), S29--S36 (2019).
\newblock \doi{10.1080/0284186x.2019.1584404}

\bibitem{Bodla2017}
Bodla, N., Singh, B., Chellappa, R., Davis, L.S.: Soft-{NMS} {\textemdash} improving object detection with one line of code.
\newblock In: 2017 {IEEE} International Conference on Computer Vision ({ICCV}). {IEEE} (2017).
\newblock \doi{10.1109/iccv.2017.593}

\bibitem{Center2009}
Center, M.M., Jemal, A., Ward, E.: International trends in colorectal cancer incidence rates.
\newblock Cancer Epidemiology Biomarkers {\&} Prevention \textbf{18}(6), 1688--1694 (2009).
\newblock \doi{10.1158/1055-9965.epi-09-0090}

\bibitem{Chen2018}
Chen, M., Du, P., Zhang, D.: Massive colonoscopy images oriented polyp detection.
\newblock In: Proceedings of the 2018 5th International Conference on Biomedical and Bioinformatics Engineering - {ICBBE} {\textquotesingle}18. {ACM} Press (2018).
\newblock \doi{10.1145/3301879.3301903}

\bibitem{Fauzi2019}
Fauzi, M.F.A., Chen, W., Knight, D., Hampel, H., Frankel, W.L., Gurcan, M.N.: Tumor budding detection system in whole slide pathology images.
\newblock Journal of Medical Systems \textbf{44}(2) (2019).
\newblock \doi{10.1007/s10916-019-1515-y}

\bibitem{Girshick2015}
Girshick, R.: Fast r-{CNN}.
\newblock In: 2015 {IEEE} International Conference on Computer Vision ({ICCV}). {IEEE} (2015).
\newblock \doi{10.1109/iccv.2015.169}

\bibitem{Girshick2014}
Girshick, R., Donahue, J., Darrell, T., Malik, J.: Rich feature hierarchies for accurate object detection and semantic segmentation.
\newblock In: 2014 {IEEE} Conference on Computer Vision and Pattern Recognition. {IEEE} (2014).
\newblock \doi{10.1109/cvpr.2014.81}

\bibitem{Hwang2007}
Hwang, S., Oh, J., Tavanapong, W., Wong, J., de~Groen, P.C.: Polyp detection in colonoscopy video using elliptical shape feature.
\newblock In: 2007 {IEEE} International Conference on Image Processing. {IEEE} (2007).
\newblock \doi{10.1109/icip.2007.4379193}

\bibitem{Jia2020}
Jia, X., Mai, X., Cui, Y., Yuan, Y., Xing, X., Seo, H., Xing, L., Meng, M.Q.H.: Automatic polyp recognition in colonoscopy images using deep learning and two-stage pyramidal feature prediction.
\newblock {IEEE} Transactions on Automation Science and Engineering pp. 1--15 (2020).
\newblock \doi{10.1109/tase.2020.2964827}

\bibitem{Karkanis2003}
Karkanis, S., Iakovidis, D., Maroulis, D., Karras, D., Tzivras, M.: Computer-aided tumor detection in endoscopic video using color wavelet features.
\newblock {IEEE} Transactions on Information Technology in Biomedicine \textbf{7}(3), 141--152 (2003).
\newblock \doi{10.1109/titb.2003.813794}

\bibitem{Li2018}
Li, Z., Peng, C., Yu, G., Zhang, X., Deng, Y., Sun, J.: Detnet: A backbone network for object detection

\bibitem{Lin2016}
Lin, T.Y., Dollár, P., Girshick, R., He, K., Hariharan, B., Belongie, S.: Feature pyramid networks for object detection

\bibitem{Lin2017}
Lin, T.Y., Goyal, P., Girshick, R., He, K., Dollar, P.: Focal loss for dense object detection.
\newblock In: 2017 {IEEE} International Conference on Computer Vision ({ICCV}). {IEEE} (2017).
\newblock \doi{10.1109/iccv.2017.324}

\bibitem{Liu2016ssd}
Liu, W., Anguelov, D., Erhan, D., Szegedy, C., Reed, S., Fu, C.Y., Berg, A.C.: Ssd: Single shot multibox detector.
\newblock In: Computer Vision {\textendash} ECCV 2016, pp. 21--37. Springer International Publishing (2016).
\newblock \doi{10.1007/978-3-319-46448-0\_2}

\bibitem{Redmon2016}
Redmon, J., Divvala, S., Girshick, R., Farhadi, A.: You only look once: Unified, real-time object detection.
\newblock In: 2016 {IEEE} Conference on Computer Vision and Pattern Recognition ({CVPR}). {IEEE} (2016).
\newblock \doi{10.1109/cvpr.2016.91}

\bibitem{Ren2017}
Ren, S., He, K., Girshick, R., Sun, J.: Faster r-{CNN}: Towards real-time object detection with region proposal networks.
\newblock {IEEE} Transactions on Pattern Analysis and Machine Intelligence \textbf{39}(6), 1137--1149 (2017).
\newblock \doi{10.1109/tpami.2016.2577031}

\bibitem{Shrivastava2016}
Shrivastava, A., Gupta, A., Girshick, R.: Training region-based object detectors with online hard example mining.
\newblock In: 2016 {IEEE} Conference on Computer Vision and Pattern Recognition ({CVPR}). {IEEE} (2016).
\newblock \doi{10.1109/cvpr.2016.89}

\bibitem{Shrivastava2016a}
Shrivastava, A., Sukthankar, R., Malik, J., Gupta, A.: Beyond skip connections: Top-down modulation for object detection

\bibitem{Shussman2014}
Shussman, N., Wexner, S.D.: Colorectal polyps and polyposis syndromes.
\newblock Gastroenterology Report \textbf{2}(1), 1--15 (2014).
\newblock \doi{10.1093/gastro/got041}

\bibitem{Siegel2020}
Siegel, R.L., Miller, K.D., Sauer, A.G., Fedewa, S.A., Butterly, L.F., Anderson, J.C., Cercek, A., Smith, R.A., Jemal, A.: Colorectal cancer statistics, 2020.
\newblock {CA}: A Cancer Journal for Clinicians  (2020).
\newblock \doi{10.3322/caac.21601}

\bibitem{Storn1997}
Storn, R., Price, K.: Differential evolution – a simple and efficient heuristic for global optimization over continuous spaces.
\newblock Journal of Global Optimization \textbf{11}(4), 341--359 (1997).
\newblock \doi{10.1023/a:1008202821328}

\bibitem{Sundaram2019}
Sundaram, P.S., Santhiyakumari, N.: An enhancement of computer aided approach for colon cancer detection in {WCE} images using {ROI} based color histogram and {SVM}2.
\newblock Journal of Medical Systems \textbf{43}(2) (2019).
\newblock \doi{10.1007/s10916-018-1153-9}

\bibitem{Tajbakhsh2015}
Tajbakhsh, N., Gurudu, S.R., Liang, J.: Automatic polyp detection in colonoscopy videos using an ensemble of convolutional neural networks.
\newblock In: 2015 {IEEE} 12th International Symposium on Biomedical Imaging ({ISBI}). {IEEE} (2015).
\newblock \doi{10.1109/isbi.2015.7163821}

\bibitem{Triggsa}
Triggs, B.: Empirical filter estimation for subpixel interpolation and matching.
\newblock In: Proceedings Eighth {IEEE} International Conference on Computer Vision. {ICCV} 2001. {IEEE} Comput. Soc.
\newblock \doi{10.1109/iccv.2001.937674}

\bibitem{Uijlings2013}
Uijlings, J.R.R., van~de Sande, K.E.A., Gevers, T., Smeulders, A.W.M.: Selective search for object recognition.
\newblock International Journal of Computer Vision \textbf{104}(2), 154--171 (2013).
\newblock \doi{10.1007/s11263-013-0620-5}

\bibitem{Wang2014}
Wang, Y., Tavanapong, W., Wong, J., Oh, J., de~Groen, P.C.: Part-based multiderivative edge cross-sectional profiles for polyp detection in colonoscopy.
\newblock {IEEE} Journal of Biomedical and Health Informatics \textbf{18}(4), 1379--1389 (2014).
\newblock \doi{10.1109/jbhi.2013.2285230}

\bibitem{Yu2015}
Yu, F., Koltun, V.: Multi-scale context aggregation by dilated convolutions

\bibitem{Yu2017a}
Yu, F., Koltun, V., Funkhouser, T.: Dilated residual networks

\bibitem{Yu2017}
Yu, L., Chen, H., Dou, Q., Qin, J., Heng, P.A.: Integrating online and offline three-dimensional deep learning for automated polyp detection in colonoscopy videos.
\newblock {IEEE} Journal of Biomedical and Health Informatics \textbf{21}(1), 65--75 (2017).
\newblock \doi{10.1109/jbhi.2016.2637004}

\bibitem{Zhang2017}
Zhang, R., Zheng, Y., Mak, T.W.C., Yu, R., Wong, S.H., Lau, J.Y.W., Poon, C.C.Y.: Automatic detection and classification of colorectal polyps by transferring low-level {CNN} features from nonmedical domain.
\newblock {IEEE} Journal of Biomedical and Health Informatics \textbf{21}(1), 41--47 (2017).
\newblock \doi{10.1109/jbhi.2016.2635662}

\bibitem{Zhang2019}
Zhang, X., Chen, F., Yu, T., An, J., Huang, Z., Liu, J., Hu, W., Wang, L., Duan, H., Si, J.: Real-time gastric polyp detection using convolutional neural networks.
\newblock {PLOS} {ONE} \textbf{14}(3), e0214133 (2019).
\newblock \doi{10.1371/journal.pone.0214133}

\bibitem{Zhong2020}
Zhong, Q., Li, C., Zhang, Y., Xie, D., Yang, S., Pu, S.: Cascade region proposal and global context for deep object detection.
\newblock Neurocomputing \textbf{395}, 170--177 (2020).
\newblock \doi{10.1016/j.neucom.2017.12.070}

\bibitem{Zlocha2019a}
Zlocha, M., Dou, Q., Glocker, B.: Improving retinanet for ct lesion detection with dense masks from weak recist labels

\end{thebibliography}


\begin{thebibliography}{}
%
%
\bibitem{RefJ}
Author, Article title, Journal, Volume, page numbers (year)
\bibitem{RefB}
Author, Book title, page numbers. Publisher, place (year)
\end{thebibliography}

\end{document}